\documentclass{article}
\usepackage{ijcai25}

\usepackage{times}
\usepackage{soul}
\usepackage{url}
\usepackage[hidelinks]{hyperref}
\usepackage[utf8]{inputenc}
\usepackage[small]{caption}
\usepackage{graphicx}
\usepackage{amsmath}
\usepackage{amsthm}
\usepackage{booktabs}
\usepackage{algorithm}
\usepackage{algorithmic}
\usepackage{amssymb}
\usepackage{enumitem}
\usepackage[switch]{lineno}
\usepackage[ruled, vlined, linesnumbered, algo2e]{algorithm2e}
\usepackage[table,xcdraw]{xcolor}

\title{Rethinking Federated Graph Learning: A Data
Condensation Perspective}

\author{
Hao Zhang$^{1,2}$
\and
Xunkai Li$^3$\and
Yinlin Zhu$^4$\And
Lianglin Hu$^{1}$\\
\affiliations
$^1$ Computer Network Information Center, Chinese Academy of Sciences, Beijing\\
$^2$ University of Chinese Academy of Sciences, Beijing, China\\
$^3$ Beijing Institute of Technology, Beijing, China\\
$^4$ Sun Yat-sen University, Guangzhou, China\\
\emails
hzhang@cnic.cn,
cs.xunkai.li@gmail.com,
zhuylin27@mail2.sysu.edu.cn,
hull@cnic.cn
}

\begin{document}

\maketitle

\begin{abstract}
    Federated graph learning is a widely recognized technique that promotes collaborative training of graph neural networks (GNNs) by multi-client graphs.
    However, existing approaches heavily rely on the communication of model parameters or gradients for federated optimization and fail to adequately address the data heterogeneity introduced by intricate and diverse graph distributions.
    Although some methods attempt to share additional messages among the server and clients to improve federated convergence during communication, they introduce significant privacy risks and increase communication overhead.
    To address these issues, we introduce the concept of a condensed graph as a novel optimization carrier to address FGL data heterogeneity and propose a new FGL paradigm called FedGM. 
    Specifically, we utilize a generalized condensation graph consensus to aggregate comprehensive knowledge from distributed graphs, while minimizing communication costs and privacy risks through a single transmission of the condensed data. 
    Extensive experiments on six public datasets consistently demonstrate the superiority of FedGM over state-of-the-art baselines, highlighting its potential for a novel FGL paradigm.
\end{abstract}

\section{Introduction}

    Graph Neural Networks (GNNs) have emerged as a robust machine learning paradigm to learn expressive representations of graph-structured data through message passing, exhibiting remarkable performance across various AI applications, such as molecular interactions\cite{huang2020skipgnn}.
    However, most existing GNNs adopt a centralized training strategy where graph data need to be collected together before training.
    In practical industrial scenarios, large-scale graphs are collected and stored on edge devices. 
    Meanwhile, regulations such as GDPR~\cite{gdpr2017eu} highlight the importance of data privacy and impose restrictions on the transmission of local data, which often contains sensitive information.
    This has led to the exploration of leveraging collective intelligence through distributed data silos to enable collaboration in graph learning~\cite{li2022flondatasilos}.
    
    To this end, Federated Graph Learning (FGL) has been proposed, extending Federated Learning (FL) to graph-structured data. 
    Its core idea is to harness collective intelligence for the collaborative training of powerful GNNs, thereby advancing AI-driven insights in federated systems.
    Given the diversity of graph-based downstream tasks, this paper focuses on subgraph-FL, the instance of FGL on a semi-supervised node classification paradigm. 
    To enhance understanding, we present case study within healthcare systems.

    \textbf{Case Study.} 
    In different regions, residents visit various hospitals (e.g., independent clients), all of which are centrally managed by government organizations (e.g., the trusted server). 
    Each hospital maintains a subgraph in its database, containing demographics, living conditions, and patient interactions. 
    These subgraphs form a global patient network.
    Notably, due to privacy regulations, geographic isolation, and competitive concerns, centralized data storage is not feasible. 
    Fortunately, subgraph-FL enables federated training through multi-client collaboration without direct data sharing. 
    For tasks such as predicting the spread of infections during a pandemic~\cite{bertozzi2020challenges}, developing a federated collaborative paradigm based on distributed scenarios is essential.

\begin{figure}[t]
  \includegraphics[width=0.49\textwidth]{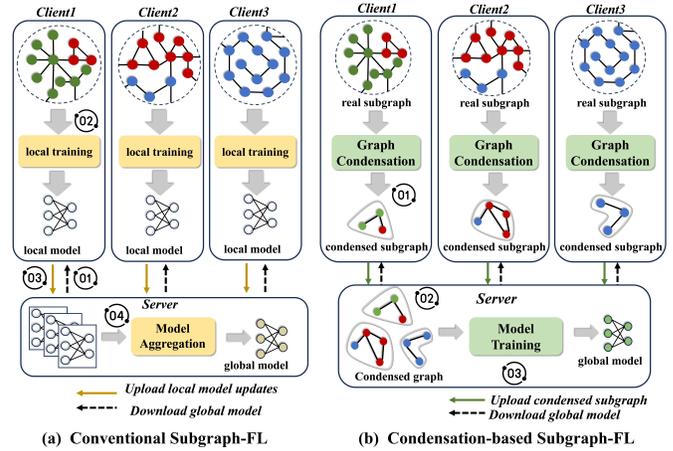}
  \captionsetup{font={small,stretch=1}}
  \caption{ 
  (a) The conventional subgraph-FL framework in the subgraph heterogeneity scenario where node colors represent different labels. 
  (b) The condensation-based subgraph-FL framework, which trains a robust global model by integrating condensed knowledge. 
  }
\label{fig: intro_fig}
\vspace{-0.6cm}
\end{figure}

    Specifically, as illustrated in Fig.\ref{fig: intro_fig}(a), the iterative training process of conventional subgraph-FL consists of four steps: (1) all clients download the latest global model from the server; 
    (2) each client trains the model on its privately stored subgraph;
    (3) after local training, the clients upload the model parameters or gradients to the central server; 
    (4) the server aggregates the model parameters or gradients to update the global model, which is then broadcast back to clients.
    
    Based on this, most subgraph-FL approaches \cite{fu2022fgl_survey} rely on model parameters or gradients as the optimization carriers for federated training. 
    These carriers are primarily derived from Computer Vision (CV)-based FL collaborative paradigm~\cite{lim2020fl_survey}, which struggles to capture the client-specific local convergence tendencies due to the complex topology in FGL, thus failing to adequately address the unique challenges of subgraph heterogeneity \cite{baek2023personalized}.
    Notably, current FGL methods \cite{li2024fedgta,li2024adafgl} typically face a trade-off between optimization and privacy, as they seek to enhance model gradient-based federated convergence by sharing more messages. 
    Further analysis and empirical studies will be provided in Sec.~\ref{sec:empirical analysis}. 
    Despite the significant contributions of existing methods, the inherent trade-off dilemma in CV-based federated optimization carriers limits the upper bound of FGL convergence, motivating us to propose a new collaborative paradigm.

    Inspired by Graph Condensation (GC)~\cite{zhao2020dc_gm,jin2021gc_gm}, a comprehensive and explicit approach based on condensed graphs holds promise in addressing the aforementioned limitations. 
    Specifically, condensed graphs can effectively capture the complex relationships between nodes and topology, providing a more suitable means of information transmission.
    This implies that we can use the condensed graph as the optimization carrier for FGL, replacing traditional model parameters or gradients.
    Furthermore, a recent study~\cite{dong2022privacy} suggests that data condensation via gradient matching can safeguard privacy. 
    This makes condensed graph-based approaches highly promising as a robust framework for FGL data heterogeneity. 
    The core idea of our method is to integrate generalized condensation subgraph consensus to acquire comprehensive and reliable knowledge.

     In this paper, we propose a new FGL paradigm based on condensed subgraphs, as illustrated in Fig.\ref{fig: intro_fig}(b): 
    (1) Each client performs local subgraph condensation using gradient matching and then uploads the condensed subgraph to the server; 
    (2) The central server optimizes the condensed knowledge from a global perspective, resulting in a global-level condensed graph;
    (3) The server then trains a robust global model on the condensed graph and returns it to the clients.

%\subsection{An Instantiation of Our Proposed Framework}

    Based on our proposed FGL paradigm, we introduce FedGM, a specialized dual-stage framework, as follows: 
\textbf{Stage 1:} 
    Each client independently performs local subgraph condensation through gradient matching between the real subgraph and the condensed subgraph, without any communication, and then uploads the condensed knowledge to the server.
    Notably, multiple clients execute data condensation locally and in parallel. 
    The central server then integrates these condensed subgraphs into a global-level condensed graph.
    It implies that Stage 1 is completed. And clients and the server only need to perform a single communication round to upload the locally condensed subgraphs, making this a one-shot FGL process.
\textbf{Stage 2:} 
    To achieve performance comparable to directly training on the implicit global real graph, FedGM employs federated gradient matching to optimize the condensed features. 
    This approach leverages global class-wise knowledge to reinforce and consolidate the condensation consensus through multiple rounds of federated optimization.
    Subsequently, the central server trains a global robust GNN using the global condensed graph and distributes it to all clients.
    
    Our contributions are as follows:
    (1) \textbf{New Framework}. 
    To the best of our knowledge, we are the first to introduce condensed graphs as a novel optimization carrier to address the challenge of subgraph heterogeneity.
    (2) \textbf{New Paradigm}. 
    We propose FedGM, a dual-stage paradigm that integrates generalized condensed subgraph consensus to obtain comprehensive knowledge while minimizing communication costs and reducing the risk of privacy breaches through a single transmission of condensed data between clients and the server.
    (3) \textbf{SOTA Performance}. 
    Extensive experiments on six datasets demonstrate the consistent superiority of FedGM over state-of-the-art baselines, with improvements of up to 4.3\%.

\section{Notations and Problem Formalization}

%NOTATIONS AND PROBLEM FORMALIZATION
\subsection{Notations}

\textbf{Graph Neural Networks.}
    Consider a graph $G = \{\mathbf{A},\mathbf{X},\mathbf{Y}\}$ consisting of $N$ nodes, where $\mathbf{X} \in \mathbb{R}^{N\times d}$ is the $d$-dimensional node feature matrix and $\mathbf{Y} \in \{1,...,C\}^N$ denotes the node labels over $C$ classes. $\mathbf{A}\in \mathbb{R}^{N\times N}$ is the adjacency matrix, with entry $\mathbf{A}_{i,j} > 0$ denoting an observed edge from node $i$ to $j$, and $\mathbf{A}_{i,j} = 0$ otherwise. 
    Building upon this, most GNNs can be subsumed into the deep message-passing framework~\cite{wu2020gnn_survey}. 
    We use graph convolutional network (GCN)~\cite{zhang2019gcn} as an example, where the propagation process in the $\ell$-th layer is as follows:
    
\begin{equation}
\mathbf{H}^{(\ell)}=\text{ReLU}\left(\hat{\mathbf{A}}\mathbf{H}^{(\ell-1)}\mathbf{W}^{(\ell)}\right),
\end{equation}

\noindent
    where $\hat{\mathbf{A}}=\tilde{\mathbf{D}}^{-\frac12}\tilde{\mathbf{A}}\tilde{\mathbf{D}}^{-\frac12}$ is the normalized adjacency matrix. $\tilde{\mathbf{A}}$ is the adjacency matrix with the self-loop, $\tilde{\mathbf{D}}$ is the degree matrix and $\mathbf{W}^{(\ell)}$ is the trainable weights at layer $\ell$. $\mathbf{H}^{(\ell)}$ is the output node embeddings from the $\ell$-th layer.
    
\vspace{1mm} 

\noindent\textbf{Graph Condensation.}
    Graph condensation is proposed to learn a synthetic graph with $N'\ll N$ nodes from the real graph $G$, denoted by $\mathcal{S}_k=\{\mathbf{A}',\mathbf{X}',\mathbf{Y}'\}$ with $\mathbf{A}'\in{\mathbb{R}^{N'\times N'}}$, $\mathbf{X}'\in\mathbb{R}^{N'\times d}$, $\mathbf{Y}'\in\{1,...,C\}^{N'}$, such that a GNN $f(\cdot)$ solely trained on $\mathcal{S}$ can achieve comparable performance to the one trained on the original graph. In other words, graph condensation can be considered as a process of minimizing the loss defined on the models trained on the real graph $G$ and the synthetic graph $\mathcal{S}$:
    
 \begin{equation}
 \mathcal{S} = \underset{\mathcal{S}}{\text{arg}\min}\mathcal{L}(\text{GNN}_{\theta_\mathcal{S}}(G),\text{GNN}_{\theta_G}(G)),
 \end{equation}

 \noindent
    where $\text{GNN}_{\theta_{\mathcal{S}}}$ and $\text{GNN}_{\theta_{G}}$ denote the GNN models trained on $\mathcal{S}$ and $G$, respectively; $\mathcal{L}$ represents the loss function used to measure the difference of these two models.
    
\vspace{1mm} 

\noindent\textbf{Subgraph Federated Learning.}
    In subgraph-FL, the $k$-th client has a subgraph $G_k = \{\mathbf{A}_k,\mathbf{X}_k,\mathbf{Y}_k\}$ of an implicit global graph $G_{glo} = \{\mathbf{A}_{glo},\mathbf{X}_{glo},\mathbf{Y}_{glo}\}$ (i.e., $\mathbf{A}_k\subseteq \mathbf{A}_{glo}, \mathbf{X}_k\subseteq \mathbf{X}_{glo}, \mathbf{Y}_k\subseteq \mathbf{Y}_{glo}$). Each subgraph consists of $N_k$ nodes, where $\mathbf{X}_k \in \mathbb{R}^{N_k\times d}$ is the $d$-dimensional node feature matrix and $\mathbf{Y_k} \in \{1,...,C\}^{N_k}$ denotes the node labels over $C$ classes. Typically, the training process for the $t$ -th communication round in subgraph-FL with the FedAvg aggregation can be described as follows: 
    (i) \textit{Initialization:} This step occurs only at the first communication round ($t=1$). The server sets the local GNN parameters of $k$ clients to the global GNN parameters $\Bar{\mathbf{\theta}}$, using $\mathbf{\theta}_{k} \leftarrow \Bar{\mathbf{\theta}}\, \forall k$.
    (ii) \textit{Local Updates:} Each local GNN performs training on the local data $G_k$ to minimize the task loss $\mathcal{L}(G_k;\theta_k)$, and then updating the parameters: 
$\mathbf{\theta}_k \leftarrow \mathbf{\theta}_k - \eta \nabla \mathcal{L}$.
    (iii) \textit{Global Aggregation:} After local training, the server aggregates local knowledge with respect to the number of training instances, i.e., $\Bar{\mathbf{\theta}} \leftarrow \frac{N_k}{N} \sum_{k=1}^K \mathbf{\theta}_k$ with $N = \sum_k N_k$, and distributes the updated global parameters $\bar{\mathbf{\theta}}$ to clients selected at the next round.

\subsection{Problem Formalization}
\label{subsec: problem}
    The proposed condensation-based Subgraph-FL framework is as follows: Firstly, each client performs local subgraph condensation, then uploads condensed knowledge to the server. Specifically, suppose that a client $k$ is tasked with learning a local condensed subgraph with $N'< N$ nodes from the real subgraph $G_k$, denoted as $\mathcal{S}_k=\{\mathbf{A}'_k,\mathbf{X}'_k,\mathbf{Y}'_k\}$ with $\mathbf{A}_k'\in{\mathbb{R}^{N'\times N'}}$, $\mathbf{X}_k'\in\mathbb{R}^{N'\times d}$, $\mathbf{Y}_k'\in\{1,...,C\}^{N'}$.
    The central server leverages the global perspective to optimize the condensed subgraphs, resulting in a global-level condensed graph $\mathcal{S}_{glo} = \{\mathbf{A}'_{glo},\mathbf{X}'_{glo},\mathbf{Y}'_{glo}\}$.
    Subsequently, the server trains a robust global model on the condensed graph and returns it to the clients.
    The motivation for this design is to obtain a powerful model trained on the condensed graph $\mathcal{S}_{glo}$, achieving performance comparable to one directly trained on the implicit global graph $G_{glo}$:
\begin{equation}
\begin{aligned}
\label{eq: global gc}
    &\min\limits_\mathcal{S}\mathcal{L}(\text{GNN}_{\theta_\mathcal{S}}(\mathbf{A}_{glo},\mathbf{X}_{glo}),\mathbf{Y}_{glo})\\ &\text{s.t}\ \theta_\mathcal{S}= \underset{\theta}{\text{arg}\min}\mathcal{L}(\text{GNN}_\theta(\mathbf{A}'_{glo},\mathbf{X}'_{glo}),\mathbf{Y}'_{glo})
\end{aligned}
\end{equation}
    where $\text{GNN}_{\theta}$ denotes the GNN model parameterized with $\theta$, $\theta_{\mathcal{S}}$ denotes the parameters of the model trained on $\mathcal{S}_{glo}$, and $\mathcal{L}$ is the loss function used to measure the difference between model predictions and ground truth (i.e. cross-entropy loss).

\vspace{0.4cm}

\section{Empirical Analysis}
\label{sec:empirical analysis}

\begin{figure}[t]
  \includegraphics[width=0.49\textwidth]{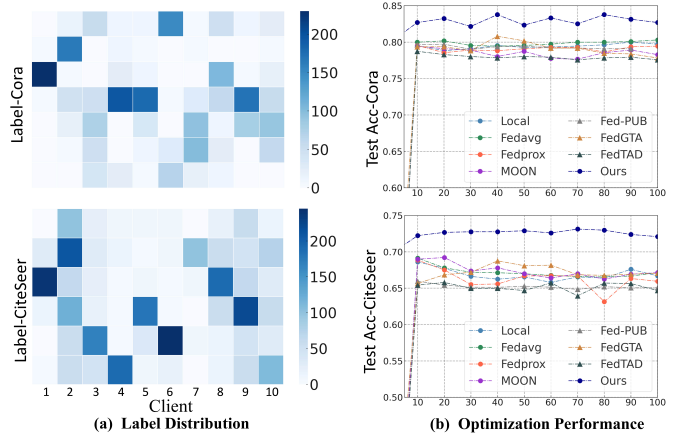}
  \captionsetup{font={small,stretch=1}}
  \caption{ 
  (a) Label distribution based on random data split, where the color gradient from white to blue indicates the increasing number of nodes held by different clients in each class. (b) Optimization performance of various methods under subgraph heterogeneity scenarios. The x-axis of the line plot represents federated training rounds, with "Local" indicating model performance in siloed settings.
  }
\label{fig: sec3}
\vspace{-0.2cm}
\end{figure}

%\subsection{FGL Optimization Dilemma}
    In this section, we empirically explore subgraph heterogeneity to investigate the FGL optimization dilemma. According to our observations, existing methods have the following limitations:
    (i) Using model parameters or gradients as primary information carriers overlooks the complex interplay between features and topology within local heterogeneous subgraphs, resulting in sub-optimal performance;
    (ii) many existing methods require the upload of additional information (e.g., subgraph embeddings, or mixed moments), raising privacy concerns. The in-depth analysis is provided as follows.
    
    In CV-based FL, data heterogeneity refers to variations among clients in terms of features, labels, data quality, and data quantity~\cite{qu2022dh,li2022dh_2}.
    This variability presents substantial challenges for effective federated training and optimization. 
    In this work, we focus on the heterogeneity of features and labels, considering their strong correlation, widely highlighted by practical applications.
    Unlike data heterogeneity in conventional FL, subgraph heterogeneity is influenced by diverse topologies across clients~\cite{li2024adafgl}.
    According to the homophily assumption~\cite{wu2020comprehensive}, connected nodes tend to share similar feature distributions and labels.
    However, with the increasing deployment of GNNs in real-world applications, topology heterophily has emerged~\cite{zhu2021heterophily_1,luan2022heterophily_2}, where the connected nodes exhibit contrasting attributes.
    Due to community-based subgraph data collection methods (i.e., the community can be viewed as the client), subgraphs from different communities often exhibit diverse topological structures, leading to Non-independent and identically distributed labels, as shown in Fig.\ref{fig: sec3}(a). 
    Specifically, we observe strong homophily at Client 1 in Cora and CiteSeer, as the majority of nodes belong to the same label class. In contrast, at Cora-Client 3 and CiteSeer-Client 10, we observe the presence of heterophily, as label distributions approach uniformity.
    In summary, clients often exhibit diverse label distributions and topologies, a characteristic of distributed graphs that demands attention. 
    Conventional federated training, which neglects subgraph heterogeneity, leads to underperformance.

\begin{figure*}[t]
  \includegraphics[width=0.998\textwidth]{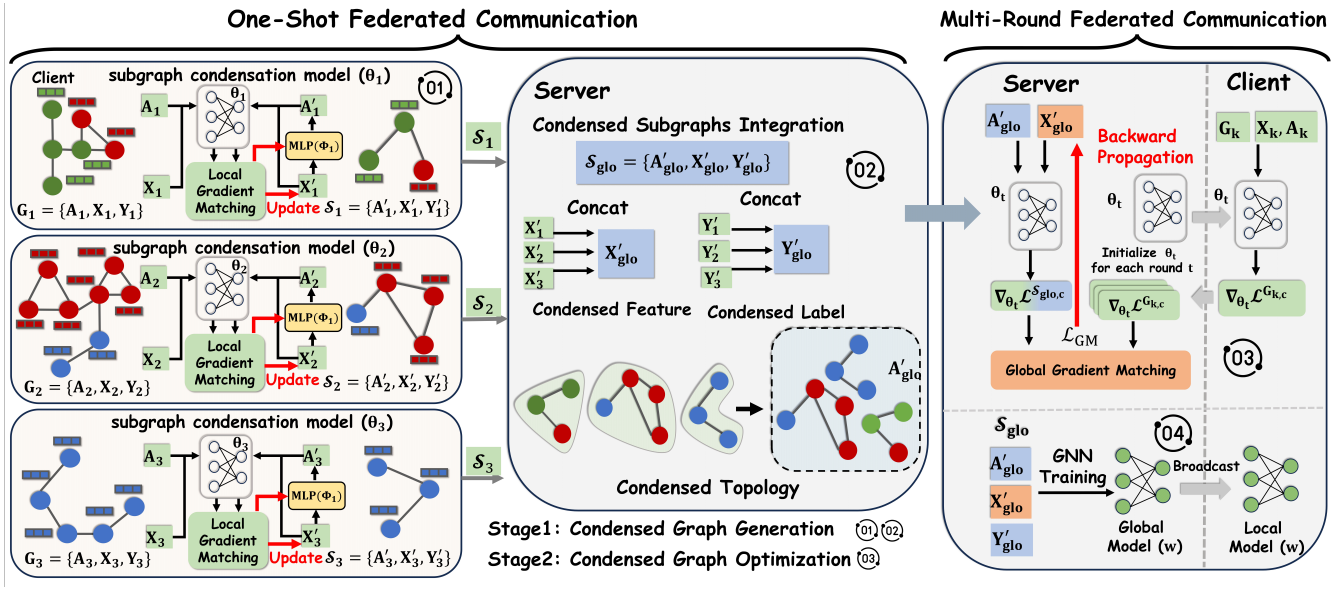}
  \vspace{-0.6cm}
  \captionsetup{font={small,stretch=1}}
  \caption{Overview of our proposed FedGM paradigm. We first perform local subgraph condensation and the central server integrates the condensed subgraphs. Subsequently, the server receives class-wise gradients from real subgraphs in federated communication to enhance the quality of condensed knowledge. Ultimately, the global model is trained on the condensed graph and then distributed to the clients.}
\label{fig: framwork}
\vspace{-0.5cm}
\end{figure*}

    To address these issues, FED-PUB~\cite{baek2023personalized} measures subgraph similarity by transmitting subgraph embeddings for personalized aggregation.
    FedGTA~\cite{li2024fedgta} shares mixed moments and local smoothing confidence for topology-aware aggregation. 
    FedSage+\cite{zhang2021subgraph} and FedGNN\cite{wu2021fedgnn} aim to reconstruct potentially missing edges among clients, thereby aligning local objectives at the data level. 
    FedTAD~\cite{zhu2024fedtad} introduces topology-aware, data-free knowledge distillation.
    Despite the considerable efforts of subgraph heterogeneity, these methods rely on uploading additional information, leading to privacy concerns and higher communication overhead.

    Considering the capability of condensed graphs to capture complex node-to-topology relationships while preserving privacy, we propose the condensation-based subgraph-FL framework. To validate the effectiveness of condensed knowledge as an optimization carrier, we conduct experiments using GCN in a federated learning setting with 10 clients across two common datasets, Cora~\cite{kipf2016dataset} and CiteSeer~\cite{kipf2016dataset}, as shown in Fig.\ref{fig: sec3}(b). 
    The results demonstrate improved model performance and faster convergence of the condensation-based subgraph-FL method. Superior performance indicates a better capability to tackle subgraph heterogeneity, while faster convergence translates into reduced communication costs.

\section{Method}

    The overview of our proposed FedGM is depicted in Fig.\ref{fig: framwork}. 
    In \underline{\textit{Stage 1}}, each client involves a local process of standard graph condensation by one-step gradient matching and uploads condensed subgraphs to the central server. The server subsequently integrates these into a global-level graph.
    In \underline{\textit{Stage 2}}, we introduce federated optimization and perform multiple rounds of communication to leverage the class-wise knowledge, enhancing the quality of the condensed features.
    
\vspace{-1mm} 
\subsection{Stage 1: Condensed Graph Generation}
\label{sec: Condensed Graph Generation}
 \vspace{-1mm}
 
    In the first stage, our task is to integrate the condensation consensus from clients to generate a global condensed graph.
    Since the real graph is distributed among multiple participants, direct access to the real graph to obtain the condensed graph, as in Eq.(\ref{eq: global gc}), is prohibited.
    Therefore, we perform subgraph condensation on each client to achieve local optimization and then integrate these subgraphs on the server via a single round of federated communication.
    Considering privacy protection requirements and condensation quality, we adopt the gradient alignment advocated by ~\cite{jin2021gc_gm,jin2022doscond} as the local learning task. Unlike GraphGAN~\cite{wang2018graphgan} and GraphVAE~\cite{simonovsky2018graphvae}, which synthesize high-fidelity graphs by capturing data distribution, its goal is to generate informative graphs for training GNNs rather than “real-looking” graphs.
    
\begin{figure}[t]
  \includegraphics[width=0.49\textwidth]{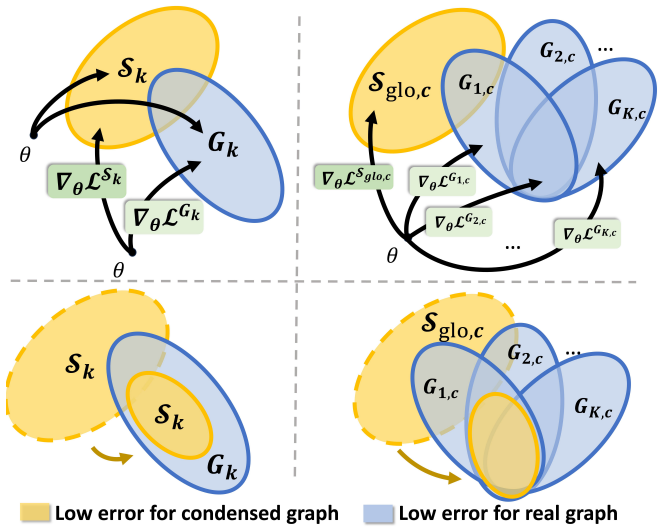}
  \vspace{-0.6cm}
  \captionsetup{font={small,stretch=1}}
  \caption{This is the representation of the local and global gradient matching in the model parameter space. The gradient matching iteratively optimizes the condensed data by minimizing the distance between gradients generated by the real and condensed data on the model, ultimately aligning the low-loss region of the condensed data within the low-loss region of the real data. The blue intersecting region in the right panel represents shared intra-class knowledge.}
  \label{fig: method_fig}
  \vspace{-0.4cm}
\end{figure}

\vspace{0.5mm}     
    \noindent\textbf{Client-Side Subgraph Condensation.}
    The FedGM aims to provide a flexible graph learning paradigm by enabling each client to perform local subgraph condensation under local conditions without requiring real-time synchronization. The client generate the condensed subgraph through one-step gradient matching\cite{jin2021gc_gm,jin2022doscond}, where a GNN is updated using real subgraph and condensed subgraph, respectively, and their resultant gradients are encouraged to be consistent, as show on the left side of Fig.\ref{fig: method_fig}. The local optimization objective can be formulated as:
\vspace{-1mm} 
\begin{equation}
\begin{aligned}
\min\limits_{\mathcal{S}_k}E_{\mathbf{\theta_k}\sim P_{\theta_k}}[D(\bigtriangledown_{\theta_k}\mathcal{L}_1,\bigtriangledown_{\theta_k}\mathcal{L}_2)],
\end{aligned}
\end{equation}
\begin{equation}
    \begin{aligned}
        \mathcal{L}_1 = \mathcal{L}(GNN_{\theta_k}(\mathbf{A}'_k,\mathbf{X}'_k), \mathbf{Y}'_k),
    \end{aligned}
\end{equation}
\begin{equation}
    \begin{aligned}
        \mathcal{L}_2 = \mathcal{L}(GNN_{\theta_k}(\mathbf{A}_k,\mathbf{X}_k), \mathbf{Y}_k),
    \end{aligned}
\end{equation}
\vspace{-1mm} 
    where $D(\cdot,\cdot)$ represents a distance function, and the subgraph condensation model parameters $\theta_k$ for client $k$ are initialized from the distribution of random initialization $P_{\theta_k}$. In each condensation round, each client initializes its subgraph condensation model to calculate the gradients for the real subgraph and the condensed subgraph. By taking different parameter initializations drawn from the distribution $P_{\theta_k}$, the learned $\mathcal{S}_k$ can avoid over fitting a specific initialization.

    To facilitate efficient learning of $\mathcal{S}_k$, a common practice is to reduce the trainable pieces in the condensed subgraph $\mathcal{S}_k = \{\mathbf{A}'_k, \mathbf{X}'_k,\mathbf{Y}'_k\}$ to only the node features $\mathbf{X}'_k$. Concretely, the labels $\mathbf{Y}'_k$ can be predefined to match the distribution of different classes in the real subgraph, while each client condenses the graph structure by leveraging a function to parameterize the adjacency matrix $\mathbf{A}'_k$ to prevent overlooking the implicit correlations between condensed node features and condensed structure~\cite{jin2021gc_gm}:
\begin{equation}
    \begin{aligned}
    \label{eq: MLP}
    \mathbf{A}'_{ij}=\sigma([\text{MLP}_\Phi([\text{x}'_{i};\text{x}'_{j}])+\text{MLP}_\Phi([\text{x}'_{j};\text{x}'_{i}])]/2),
    \end{aligned}
\end{equation}
    where $\text{MLP}_\Phi$ is a multi-layer perceptron (MLP) parameterized with $\Phi$, $[\cdot;\cdot]$ indicates concatenation and $\sigma$ is the sigmoid function. The client condenses the subgraph by optimizing alternately $\mathbf{X}'_k$ and $\Phi_k$. After condensation, the client transfers $\mathcal{S}_k$ to the server via one-shot federated communication.

\vspace{0.5mm}     
\noindent\textbf{Server-Side Condensed Subgraphs Integration.}
    To construct a global condensed graph, the server concatenates the features and labels from each client's condensed subgraph:
    \begin{equation}
        \begin{aligned}
            \mathbf{X}'_{\text{glo}} = \begin{bmatrix} \mathbf{X}'_1 \\ \mathbf{X}'_2 \\ \vdots \\ \mathbf{X}'_K \end{bmatrix}, \quad \mathbf{Y}'_{\text{glo}} = \begin{bmatrix} \mathbf{Y}'_1 \\ \mathbf{Y}'_2 \\ \vdots \\ \mathbf{Y}'_K \end{bmatrix},
        \end{aligned}
    \end{equation}
    where $\mathbf{X}'_k$ and $\mathbf{Y}'_k$ represent the condened features and labels from client $k$, respectively. Unlike real-world graph structures, the condensed topology lacks tangible meaning, which is only relevant to the passage of condensed knowledge within the GNNs.
    To avoid disrupting the knowledge representing each client's subgraph within condensed data, we retain the topology of each condensed subgraph. 
    Specifically, the global condensed adjacency matrix $\mathbf{A}'_{\text{glo}}$ is represented as:
    \begin{equation}
    \mathbf{A}'_{\text{glo}}[i,j]=\left\{\begin{aligned}
    &\mathbf{A}'_k[i,j],\ \text{if}\ i,j\in \mathcal V_k; \\
    &0,\text{otherwise,} &
    \end{aligned}\right.
    \end{equation}
    where $\mathcal V_k$ denotes the set of nodes from client $k$. Consequently, we obtain an initial global condensed graph, consisting of multiple connected components. 
    However, there remains a significant gap between the quality of the condensed graph and our desired target due to the limitations of the narrow local scope. 
    Therefore, it is crucial to effectively optimize the condensed graph by leveraging a global perspective.

\subsection{Stage 2: Condensed Graph Optimization}
\label{sec: global condensed feature optimization}
\vspace{-1mm}

    There is a common phenomenon of class imbalance at the client level in scenarios with subgraph heterogeneity, which obviously leads to poorer feature quality for condensed nodes, especially for the minority classes. 
    We observe that majority classes on one client often correspond to minority classes on other clients in scenarios with subgraph heterogeneity, as shown in Fig.~\ref{fig: intro_fig}(b).
    Therefore, we aim to collect the class-wise gradients generated by each subgraph and perform federated gradient matching to optimize the condensed features. 
    Our intuition is that the shared intra-class knowledge between clients provides a basis based on the global perspective, which iteratively reduces the gap between the condensed graph and the real graph via class-wise gradient matching, as illustrated on the right side of Fig.~\ref{fig: method_fig}.

    In the second stage, FedGM introduces condensed graph optimization, which is performed over multi-round federated communication.
    In the \( t \)-th iteration, the server samples the gradient generation model parameters  \( \theta_t \) from random initialization distribution \( P_{\theta_t} \) and sends it to each client. 

\vspace{0.5mm}     
\noindent\textbf{Clinet-Side Gradient Generation.}
    On each client, the real subgraphs generate class-wise gradients through the gradient generation model. For class $c$, and the generated gradient is:
    \begin{equation}
        \begin{aligned}\label{eq:get_local_gradient}
            \bigtriangledown_{\theta_t}\mathcal{L}^{G_{k,c}} = \bigtriangledown_{\theta_t}\mathcal{L}(GNN_{\theta_t}(\mathbf{A}_{k,c},\mathbf{X}_{k,c}), \mathbf{Y}_c),
        \end{aligned}
    \end{equation}
    where $\mathbf{A}_{k,c}$ denote the adjacency matrix composed of the \( c \)-class nodes and their neighbors and $\mathbf{X}_{k,c}$ denote the features corresponding to the \( c \)-class nodes.  To ease the presentation, we adopt the following nations for every client $k$:
    \begin{equation}
        \bigtriangledown_{\theta_t}\mathcal{L}^{G_{k}} = [\bigtriangledown_{\theta_t}\mathcal{L}^{G_{k,1}},\bigtriangledown_{\theta_t}\mathcal{L}^{G_{k,2}},...,\bigtriangledown_{\theta_t}\mathcal{L}^{G_{k,C}}].
    \end{equation}    
    \noindent Upon having the class-wise gradients obtained, clients share the generated gradients $\bigtriangledown_{\theta_t}\mathcal{L}^{G_{k}}$ with the server. 

\vspace{0.5mm}     
\noindent\textbf{Sever-Side Gradient Matching.}
    For class $c$, the server calculates the gradient generated by the implicit global graph based on the number of \( c \)-class condensed nodes from clients:
    
    \vspace{-1mm}
    \begin{equation}
        \begin{aligned}\label{eq:avg_gradient}
            \bigtriangledown_{\theta_t}\mathcal{L}^{G_c} = \sum\limits_{k=1}^K \frac{N'_{k,c}}{N'_c}\bigtriangledown_{\theta_t}\mathcal{L}^{G_{k,c}}.
        \end{aligned}
    \end{equation}
    \vspace{-2mm}
    
\noindent The gradients generated by the condensed graph is as follows:
    \begin{equation}
        \begin{aligned}\label{eq:get_glo_gradient}
        \bigtriangledown_{\theta_t}\mathcal{L}^{\mathcal{S}_{glo,c}} = \bigtriangledown_{\theta_t}\mathcal{L}(GNN_{\theta_t}(\mathbf{A}'_{glo},\mathbf{X}'_{glo}), \mathbf{Y}'_c),
        \end{aligned}
    \end{equation}
    Then we empower the server to match the gradients of the implicit global graph and the condensed graph for each category using the distance function. And we simplify our objective as
    \begin{equation}
        \begin{aligned}\label{eq: dis}
            \min\limits_{\textbf{X}'_{\text{glo}}}E_{\mathbf{\theta_t}\sim P_{\theta_t}}[D(\bigtriangledown_{\theta_t}\mathcal{L}^{G},\bigtriangledown_{\theta_t}\mathcal{L}^{\mathcal{S}_{\text{glo}}})],
        \end{aligned}
    \end{equation}
    \vspace{-2.5mm}
    
    $\bigtriangledown_{\theta_t}\mathcal{L}^{G}$ and $\bigtriangledown_{\theta_t}\mathcal{L}^{\mathcal{S}_{\text{glo}}}$ in the above equation are defined as 
    \begin{tiny}
    \begin{equation}
        \begin{aligned}
            &\bigtriangledown_{\theta_t}\mathcal{L}^{G} = [ \bigtriangledown_{\theta_t}\mathcal{L}^{G_1}, \bigtriangledown_{\theta_t}\mathcal{L}^{G_2},..., \bigtriangledown_{\theta_t}\mathcal{L}^{G_C}],\\
            &\bigtriangledown_{\theta_t}\mathcal{L}^{\mathcal{S}_{\text{glo}}} = [\bigtriangledown_{\theta_t}\mathcal{L}^{\mathcal{S}_{glo,1}},\bigtriangledown_{\theta_t}\mathcal{L}^{\mathcal{S}_{glo,2}},...,\bigtriangledown_{\theta_t}\mathcal{L}^{\mathcal{S}_{glo,C}}].&
        \end{aligned}
    \end{equation}
    \end{tiny}
    In the multi-round communication process, the central server optimizes the features using the class-specific gradients, ultimately obtaining the desired condensed graph. The second-stage method of FedGM is presented in Algorithm \ref{alg: stage3}. After federated optimization, the server trains the global model on the condensed graph and sends the final model back to clients.

 \begin{algorithm}[H]
            \caption{FedGM-Condensed Graph Optimization}
            \label{alg: stage3}
            \KwIn{Rounds, $T$; Local real subgraphs, $\{G_k\}_{k=1}^K$; Initial condensed graph, $\mathcal{S}_{glo}$}
            \KwOut{Optimized condensed graph, $\mathcal{S}_{glo}'$}
            
            \tcc{Client Execution}
            \For{ \textup{each communication round} $t = 1,...,T$}{
                \textup{Update the gradient generation model} $\theta_t$;
                \For{ \textup{each class} $c = 1,...,C$}{
                    \textup{Sample on the real subgraph according to the class} $c$  $(A_{k,c},X_{k,c},Y_{k,c})\sim G_k$;
                    
                   \textup{Calculate loss and get the gradient via Eq.\ref{eq:get_local_gradient} }
                }
                \textup{Upload the number of samples of each category and the corresponding gradient to the central server}
                
            }
            
            \tcc{Server Execution}
            \For{ \textup{each communication round} $t = 1,...,T$}{
                \textup{initialize} $\theta_t\sim P_{\theta_t}$;
                
                \For{\textup{each client} $k = 1,...,K$}{
                     \textup{Send the gradient generation model} $\theta_t$ \textup{to client k};

                     \textup{Receive the number of class samples and the corresponding gradient}
                }
                
                \For{ \textup{each class} $c = 1,...,C$}{
                    \textup{Calculate the condensed graph  gradient via Eq.\ref{eq:get_glo_gradient}};

                    \textup{Calculate the real graph gradient via Eq.\ref{eq:avg_gradient}};

                }
                \textup{Update the condensed features} $\textbf{X}'_{glo}$ \textup{via Eq.\ref{eq: dis}};

            }
    \end{algorithm}

\section{Experiments}

    In this section, we conduct experiments to verify the effectiveness of FedGM. We introduce 6 benchmark graph datasets across 5 domains and the simulation strategy for the subgraph-FL scenario. 
    And we present 8 evaluated state-of-the-art baselines.
    Specifically, we aim to answer the following questions: 
    \textbf{Q1}: Compared with other state-of-the-art federated optimization strategies, can FedGM achieve better performance?
    \textbf{Q2}: Where does the performance gain of FedGM come from? 
    \textbf{Q3}: Is FedGM sensitive to the hyperparameters?
    \textbf{Q4}: What is the time complexity of FedGM?

\subsection{Datasets and Simulation Method}

    We evaluate FedGM on six public benchmark graph datasets across five domains, including two citation networks (Cora, Citeseer) \cite{kipf2016dataset}, one co-authorship network (CS) \cite{shchur2018CS}, one co-purchase network (Amazon Photo), one task interaction network (Tolokers) \cite{platonov2023tolokers}, and one social network (Actor) \cite{actor}. 
    More details can be found in Table \ref{tab:datasets}. 
    To simulate the distributed subgraphs in subgraph-FL, we employ the Louvain algorithm \cite{blondel2008louvain} to achieve graph partitioning across 10 clients, which is based on modularity optimization and widely used in the subgraph-FL fields. 
    %\cite{zhu2024fedtad}.

\begin{table}
    \centering
    \begin{tabular}{lcccc}
        \toprule
        Dataset  & \#Nodes & \#Features & \#Edges & \#Classes \\
        \midrule
        Cora & 2,708 & 1,433 & 5,429  & 7       \\
        CiteSeer & 3,327 & 3,703 & 4,732  & 6   \\
        Photo & 7,487 & 745 & 119,043  & 8     \\
        Actor & 7,600 & 931 & 33,544 & 5 \\
        Tolokers & 11,758  & 10  & 519,000 & 2  \\
        CS  & 18,333  & 6,805 & 81,894  & 15    \\
        \bottomrule
    \end{tabular}
    \caption{Statistics of the six public benchmark graph datasets.}
    \label{tab:datasets}
    \vspace{-0.3cm}
\end{table}

\subsection{Baselines and Experimental Settings}

 \textbf{Baselines.} 
    We compare the proposed FedGM with four conventional FL optimization strategies (FedAvg \cite{mcmahan2017fedavg}, FedProx \cite{li2020fedprox}, SCAFFOLD \cite{karimireddy2020scaffold}, MOON \cite{li2021moon}), two personalized subgraph-FL optimization strategies (Fed-PUB\cite{baek2023personalized}, FedGTA\cite{li2024fedgta}), one subgraph-FL optimization strategy (FedTAD \cite{zhu2024fedtad}), and one subgraph-FL framework (FedSage+ \cite{zhang2021subgraph}).

 \textbf{Hyperparameters.} 
    For conventional framework, we employ a 2-layer GCN~\cite{kipf2016gcn} with 256 hidden units as the backbone for both the clients and the central server. The local training epoch is set to 3. Notably, model-specific baselines such as FedSage+~\cite{zhang2021subgraph} adhere to the custom architectures specified in their original papers. 
    In the FedGM framework, the local subgraph condensation model, gradient generation model, and the model employed for evaluation are all implemented as 2-layer GCNs with 256 hidden units, and the condensed graph structure generation model is implemented as 3-layer MLP with 128 hidden units. 
    In the first stage, the number of local condensation epochs is 1000. 
    Based on this, we perform the hyperparameter search for FedGM using the Optuna framework \cite{optuna_2019} on the ratio $r$ of condensed nodes to real nodes within the ranges of 0 to 1.
    For all methods, the learning rate for the GNN is set to 1e-2, the weight decay is set to 5e-4, and the dropout rate is set to 0.0. The federated training is conducted over 100 rounds. 
    For each experiment, we report the mean and variance results of 3 standardized training runs.

\subsection{Results and Analysis}

\begin{table*}[htbp]
    \setlength{\abovecaptionskip}{0.2cm} % 设置标题与表格之间的垂直间距
    \setlength{\arrayrulewidth}{1pt} 
    \renewcommand{\arraystretch}{1.8} % 表格行间距
    \caption{\textbf{Performance comparison of FedGM and baselines}, where the best and second results are highlighted in \textbf{bold} and \underline{underline}.}
    \footnotesize 
    \label{tab: compare baseline}
    \centering 
    \resizebox{150mm}{33mm}{ % 调整表格的{}整体宽度和{}高度
    \setlength{\tabcolsep}{3mm}{  % 调整列间距（即表格列之间的水平间距）
    \begin{tabular}{l|c|c|c|c|c|c|c}
        \toprule[1pt] % 粗细
         \textbf{Methods} & \multicolumn{1}{c|}{\textbf{Cora}} & \multicolumn{1}{c|}{\textbf{CiteSeer}} & \multicolumn{1}{c|}{\textbf{Photo}} & \multicolumn{1}{c|}{\textbf{Actor}} & \multicolumn{1}{c|}{\textbf{Tolokers}} & \multicolumn{1}{c|}{\textbf{CS}} & \multicolumn{1}{c}{\textbf{All Avg.}}\\

        \midrule[0.5pt]
        FedAvg~\cite{mcmahan2017fedavg}
        & \parbox[c]{1cm}{\centering 79.66 \\ \tiny ±0.15}
        & \parbox[c]{1cm}{\centering 73.34 \\ \tiny ±0.12} 
        & \parbox[c]{1cm}{\centering 90.24 \\ \tiny ±0.23}
        & \parbox[c]{1cm}{\centering 30.84 \\ \tiny ±0.15}
        & \parbox[c]{1cm}{\centering 77.99 \\ \tiny ±0.03}
        & \parbox[c]{1cm}{\centering 87.61 \\ \tiny ±0.08}
        & \parbox[c]{1cm}{\centering 73.28 \\ }
        \\
        
        FedProx~\cite{li2020fedprox}        
        & \parbox[c]{1cm}{\centering 80.08 \\ \tiny ±0.22}
        & \parbox[c]{1cm}{\centering 73.11 \\ \tiny ±0.71} 
        & \parbox[c]{1cm}{\centering 90.07 \\ \tiny ±0.33}
        & \parbox[c]{1cm}{\centering 30.72 \\ \tiny ±0.19}
        & \parbox[c]{1cm}{\centering 78.01 \\ \tiny ±0.02}
        & \parbox[c]{1cm}{\centering 88.42 \\ \tiny ±0.24}
        & \parbox[c]{1cm}{\centering \underline{73.40} \\ }
        \\

        SCAFFOLD ~\cite{karimireddy2020scaffold}       
        & \parbox[c]{1cm}{\centering 79.31 \\ \tiny ±0.48}
        & \parbox[c]{1cm}{\centering \underline{73.48} \\ \tiny ±0.00} 
        & \parbox[c]{1cm}{\centering 84.99 \\ \tiny ±0.92}
        & \parbox[c]{1cm}{\centering 28.90 \\ \tiny ±0.08}
        & \parbox[c]{1cm}{\centering 78.21 \\ \tiny ±0.25}
        & \parbox[c]{1cm}{\centering 86.60 \\ \tiny ±0.52}
        & \parbox[c]{1cm}{\centering 71.92 \\ }
       \\

        MOON ~\cite{li2021moon}        
        & \parbox[c]{1cm}{\centering 79.48 \\ \tiny ±0.30}
        & \parbox[c]{1cm}{\centering 73.41 \\ \tiny ±0.62} 
        & \parbox[c]{1cm}{\centering 89.43 \\ \tiny ±1.19}
        & \parbox[c]{1cm}{\centering 30.86 \\ \tiny ±0.13}
        & \parbox[c]{1cm}{\centering 77.98 \\ \tiny ±0.06}
        & \parbox[c]{1cm}{\centering 87.66 \\ \tiny ±0.05}
        & \parbox[c]{1cm}{\centering 73.14 \\ }
        \\

        \midrule[0.5pt]

        Fed-PUB~\cite{baek2023personalized}
        & \parbox[c]{1cm}{\centering 80.44 \\ \tiny ±0.25}
        & \parbox[c]{1cm}{\centering 70.92 \\ \tiny ±0.09} 
        & \parbox[c]{1cm}{\centering 90.63 \\ \tiny ±0.34}
        & \parbox[c]{1cm}{\centering 28.32 \\ \tiny ±0.47}
        & \parbox[c]{1cm}{\centering 78.20 \\ \tiny ±0.22}
        & \parbox[c]{1cm}{\centering 88.65 \\ \tiny ±0.33}
        & \parbox[c]{1cm}{\centering 72.86 \\ }\\

        FedSage+~\cite{zhang2021subgraph}      
        & \parbox[c]{1cm}{\centering 76.19 \\ \tiny ±1.03}
        & \parbox[c]{1cm}{\centering 71.78 \\ \tiny ±0.79} 
        & \parbox[c]{1cm}{\centering 90.50 \\ \tiny ±0.41}
        & \parbox[c]{1cm}{\centering 29.88 \\ \tiny ±0.67}
        & \parbox[c]{1cm}{\centering \underline{78.44} \\ \tiny ±0.65}
        & \parbox[c]{1cm}{\centering 87.76 \\ \tiny ±0.34}
        & \parbox[c]{1cm}{\centering 72.46 \\ }
       \\
        
         FedGTA~\cite{li2024fedgta}         
        & \parbox[c]{1cm}{\centering 78.35 \\ \tiny ±0.47}
        & \parbox[c]{1cm}{\centering 65.51 \\ \tiny ±0.15} 
        & \cellcolor{cyan!15}\parbox[c]{1cm}{\centering \textbf{91.17} \\ \tiny ±0.16}
        & \parbox[c]{1cm}{\centering 28.53 \\ \tiny ±0.15}
        & \parbox[c]{1cm}{\centering 78.16 \\ \tiny ±0.13}
        & \parbox[c]{1cm}{\centering 88.34 \\ \tiny ±0.08}
        & \parbox[c]{1cm}{\centering 71.65 \\ }
        \\
        
        FedTAD~\cite{zhu2024fedtad}        
        & \parbox[c]{1cm}{\centering 79.30 \\ \tiny ±0.17}
        & \parbox[c]{1cm}{\centering 73.41 \\ \tiny ±0.39} 
        & \parbox[c]{1cm}{\centering 80.17 \\ \tiny ±6.30}
        & \parbox[c]{1cm}{\centering \underline{30.93} \\ \tiny ±0.27}
        & \parbox[c]{1cm}{\centering 78.02 \\ \tiny ±0.41}
        & \parbox[c]{1cm}{\centering 87.83 \\ \tiny ±0.18}
        & \parbox[c]{1cm}{\centering 71.61 \\ }\\

        \midrule[0.5pt]
        \arrayrulecolor{black}
        \rowcolor{cyan!15}
        FedGM (Ours) 
        & \parbox[c]{1cm}{\centering \textbf{83.23} \\ \tiny \textbf{±0.13}}
        & \parbox[c]{1cm}{\centering \textbf{73.95} \\ \tiny \textbf{±0.65}}
        & \cellcolor{white}\parbox[c]{1cm}{\centering \underline{90.85} \\ \tiny \textbf{±0.22}}
        & \parbox[c]{1cm}{\centering \textbf{31.33} \\ \tiny \textbf{±0.31}}
        & \parbox[c]{1cm}{\centering \textbf{78.49} \\ \tiny \textbf{±0.39}}
        & \parbox[c]{1cm}{\centering \textbf{89.51} \\ \tiny \textbf{±0.08}}
        & \parbox[c]{1cm}{\centering \textbf{74.56} \\} 
        \\

        FedGM (w/o Stage 2)        
        & \parbox[c]{1cm}{\centering \underline{82.54} \\ \tiny ±0.28}
        & \parbox[c]{1cm}{\centering 72.62 \\ \tiny ±0.36} 
        & \parbox[c]{1cm}{\centering 84.91 \\ \tiny ±1.12}
        & \parbox[c]{1cm}{\centering 30.42 \\ \tiny ±0.65}
        & \parbox[c]{1cm}{\centering 78.07 \\ \tiny ±0.23}
        & \parbox[c]{1cm}{\centering \underline{88.97} \\ \tiny ±0.30}
        & \parbox[c]{1cm}{\centering 72.92 \\ }
        
        \\
        \bottomrule[1pt] % 粗细

    \end{tabular}
    }}
\end{table*}

\begin{figure}[t]
  \includegraphics[width=0.49\textwidth]{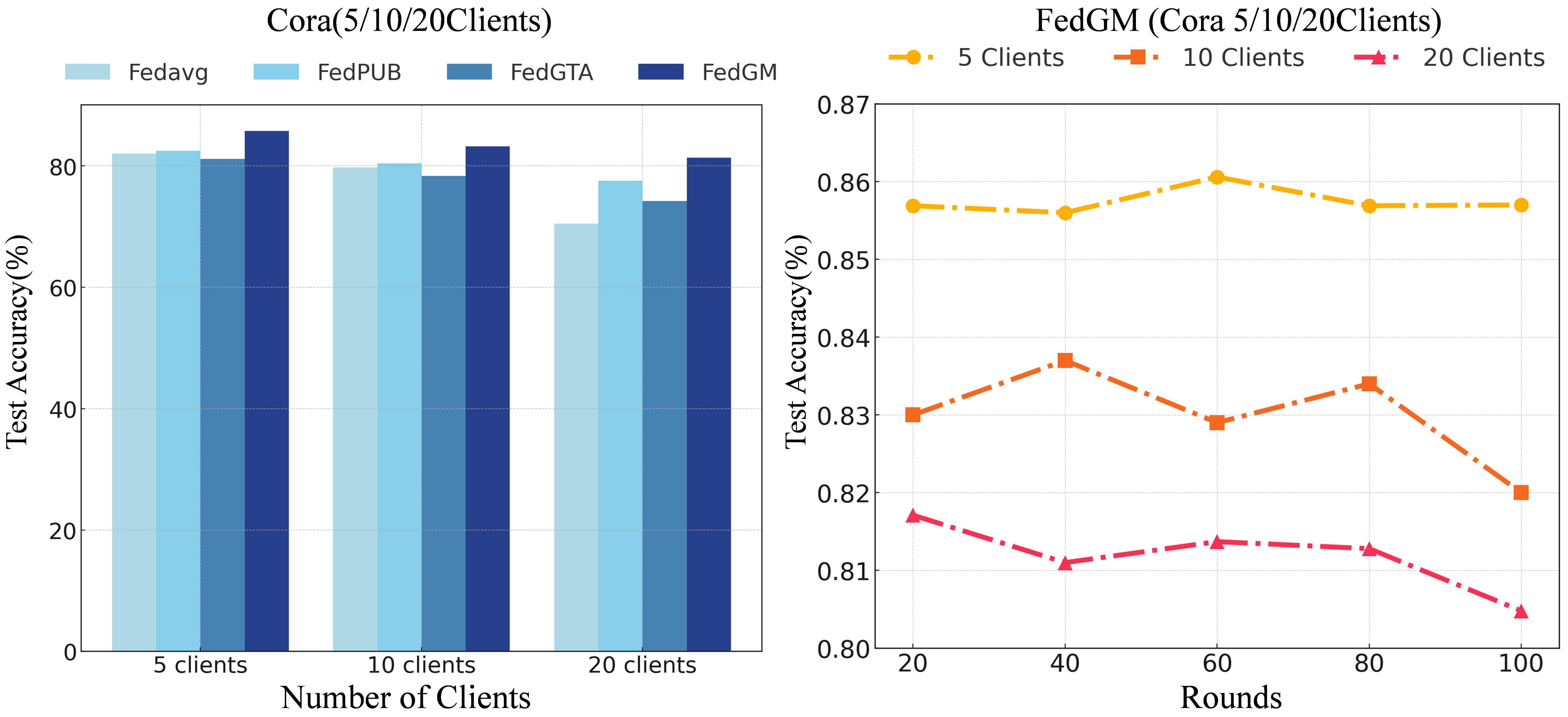}
  \vspace{-0.5cm}
  \captionsetup{font={small,stretch=1}}
  \caption{Performance of FedGM with different numbers of clients.}
  \label{fig: Clients}
  \vspace{-0.2cm}
\end{figure}

\textbf{Result 1: the answer to Q1.} The comparison results are presented in Table \ref{tab: compare baseline}. According to observations, the proposed FedGM overall outperforms the baseline. Specifically, compared with FedAvg, FedGM brings at most 4.3\%  performance improvement; Compared with Fed-PUB, FedGM can achieve a performance improvement of at most 4.1\%. Moreover, FedGM consistently outperforms existing SOTA methods in varying numbers of clients. Notably, as the number of clients increases, the performance advantage of FedGM becomes more pronounced. Specifically, with 20 clients, FedGM achieves a 13.4\% performance improvement over FedAvg, as shown in Fig.\ref{fig: Clients}. The convergence curves of FedGM and baselines are shown in Fig.\ref{fig: sec3}(b). It is observed that FedGM has a good effect at the beginning of federated communication, which shows that FedGM is suitable for subgraph-FL scenarios with limited communication overhead. In addition, FedGM demonstrates robust performance stability across various client settings, with accuracy fluctuations remaining within a margin of 2\% as shown in Fig.\ref{fig: Clients}.

\begin{figure}[t]
  \includegraphics[width=0.49\textwidth]{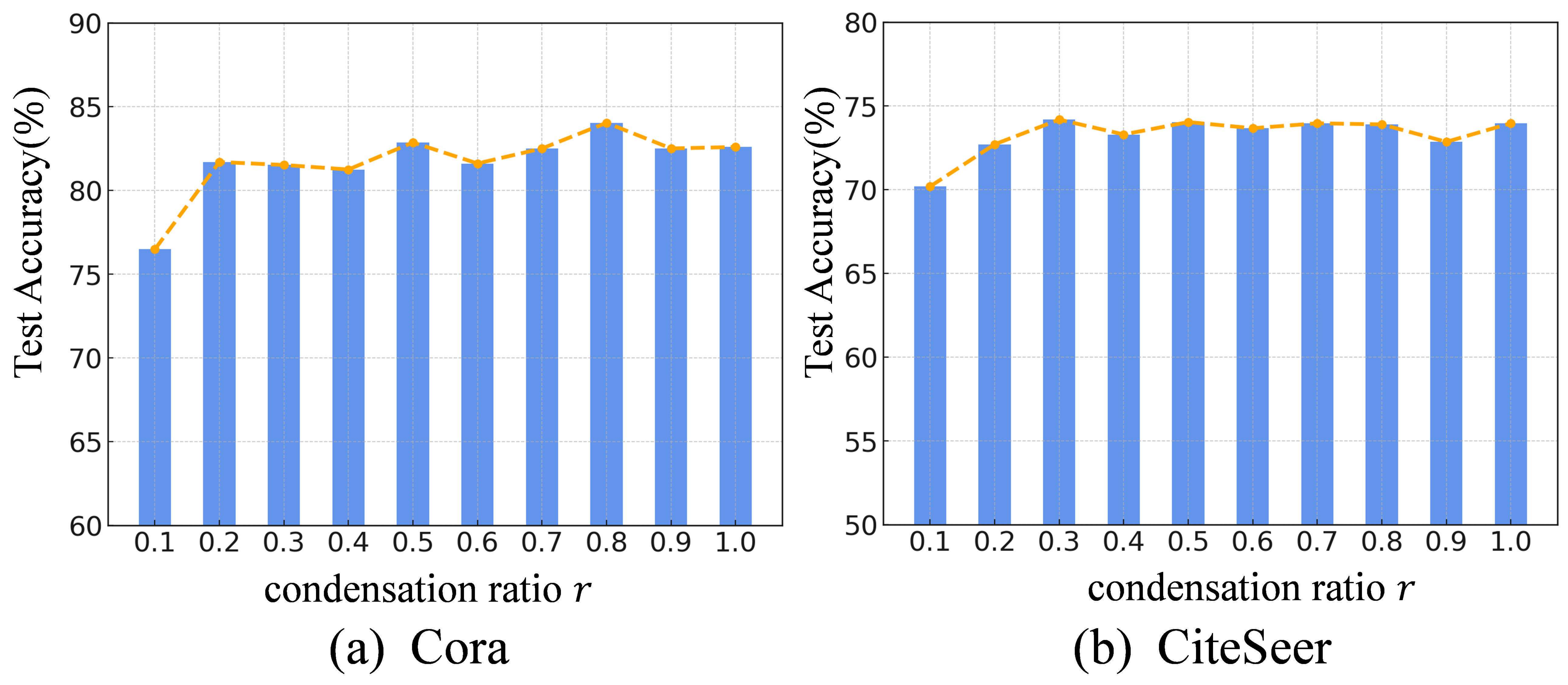}
  \vspace{-0.5cm}
  \captionsetup{font={small,stretch=1}}
  \caption{Sensitive analysis for condensation ratio $r$}
  \label{fig: Sensitive analysis}
  \vspace{-0.4cm}
\end{figure}

\textbf{Result 2: the answer to Q2.} Stage 2 builds upon the foundation established in Stage 1. To answer \textbf{Q2}, we conducted an ablation study to investigate the effectiveness of both Stage 1 and Stage 2, as shown in Table.\ref{tab: compare baseline}. After Stage 1, our method achieves an average accuracy of 72.92\%, surpassing other state-of-the-art methods on two datasets, demonstrating the feasibility of the condensation-based FGL paradigm. In addition, our method consistently achieves superior performance compared to its single-stage variant across all datasets (e.g., increasing from 84.91\% to 90.85\% and from 72.92\% to 74.56\%), highlighting the pivotal role of the second stage in enhancing the model’s representational capacity and robustness. This further validates that leveraging global intra-class knowledge contributes to improving the quality of the condensed graphs, reinforcing the efficacy of FedGM.

\textbf{Result 3: the answer to Q3.} To answer Q3, we assess the performance of FedGM under diverse condensation ratios. The sensitivity analysis on the Cora and CiteSeer datasets is presented in Fig.\ref{fig: Sensitive analysis}. Overall, most values cluster near the maximum, reflecting consistently high accuracy under the majority of conditions. FedGM is insensitive to the condensation ratio, and there is no significant dependence between performance and condensation ratio $r$.

\textbf{Result 4: the answer to Q4.} To answer \textbf{Q4}, we provide the complexity analysis of FedGM. In \underline{Stage 1}, condensation graph generation costs $\mathcal{O}(rM)$, where $r$ denotes the condensation ratio, and $M$ denotes the size of the labeled dataset. A single transmission of condensed data implies a lower risk of privacy leakage. In \underline{Stage 2}, condensation graph optimization costs $\mathcal{O}(KTN_{\Theta_{GNN}})$, where $K$ denotes the number of participating clients, $T$ denotes the number of the federal communication rounds, and $\Theta_{GNN}$ denotes the size of GNN gradients or parameters associated with gradient matching. Notably, FedAvg represents the lowest communication cost among federated learning processes, and its time complexity is also $\mathcal{O}(KTN_{\Theta_{GNN}})$. Unlike FedAvg, where the shared model parameters represent a trained GNN model, FedGM leverages the shared parameters primarily for generating gradients rather than direct model deployment. This distinction implies a reduced privacy risk during the federated process.

\section{Conclusion}
    In this paper, we conduct an in-depth analysis of the trade-off dilemma caused by the poor performance of conventional federated carriers in handling subgraph heterogeneity.
    Considering the capability of condensed graphs to capture complex node-to-topology relationships while preserving privacy, we are the first to propose a new condensation-based FGL paradigm. 
    Specifically, we propose FedGM, a dual-stage paradigm that integrates generalized condensation consensus to capture comprehensive knowledge while significantly reducing communication costs and mitigating privacy risks through a single transmission of condensed data between clients and the server.
    Experimental results demonstrate that FedGM significantly outperforms state-of-the-art baselines.

\appendix

\bibliographystyle{named}
\bibliography{ijcai25}

\end{document}